# Research Paper

# Retrieval-Augmented Generation Systems for Intellectual Property via Synthetic Multi-Angle Fine-tuning

*Runtao Ren[a], Jian Ma[a], Jianxi Luo[b]*

a Department of the Information Systems, City University of Hong Kong, Kowloon Tong, Hong Kong
b Department of Systems Engineering, City University of Hong Kong, Kowloon Tong, Hong Kong

A B S T R A C T

Retrieval-Augmented Generation (RAG) systems in the Intellectual Property (IP) field often struggle with diverse user queries, including colloquial expressions, spelling errors, and ambiguous terminology, leading to inaccurate retrieval and suboptimal responses. To address this challenge, we propose Multi-Angle Question Generation and Retrieval Fine-Tuning Method (MQG-RFM), a novel framework that leverages large language models (LLMs) to simulate varied user inquiries and fine-tunes retrieval models to align semantically equivalent but linguistically diverse questions. Unlike complex architectural modifications, MQG-RFM adopts a lightweight Data-to-Tune paradigm, combining prompt-engineered query generation with hard negative mining to enhance retrieval robustness without costly infrastructure changes. Experimental results on a Taiwan patent Q&A dataset show 185.62% improvement in retrieval accuracy on the Patent Consultation dataset and 262.26% improvement on the Novel Patent Technology Report dataset, with 14.22% and 53.58% improvements in generation quality over the baselines, respectively. By bridging the gap between user intent and system comprehension through semantic-aware retrieval optimization, MQG-RFM offers a practical, scalable approach for rapid, cost-effective deployment among small and medium-sized agencies seeking reliable patent intelligence solutions. Additionally, our proposed method has already been adopted by *ScholarMate*, the largest professional research social networking platform in China, to support real-world development and deployment. A demo version of the instantiated is available at https://github.com/renruntao/patent_rag.

\* Corresponding author.
E-mail address: runtaoren2-c@my.cityu.edu.hk(Runtao Ren), isjian@cityu.edu.hk(Jian Ma), jianxi.luo@cityu.edu.hk(Jianxi Luo)

## 1. Introduction

*The same truth can be questioned in countless ways.*

*--- Thomas Aquinas*

Half a year ago, at a patent agency in China, a lawyer tried to retrieve the legal status of a new invention from the database. However, after entering a misspelled English term, the system only returned a few irrelevant documents. Later, another user asked in a relatively complex colloquial manner, "I want to know if this device is still under review," but received a completely unrelated old patent document instead. These cases frequently appear in logs. Our further analysis of approximately 50,00 real user query records revealed that over 30% of requests involved spelling errors, colloquial expressions, or vague keywords. These examples indicate that users often pose questions in a variety of ways, as Hegel said [1]:

*The same entity can be interpreted by different languages, cultures, and backgrounds.*

Similarly, the same answer can be asked through multiple different questions. This art of language offers insights for the system design in the field of intellectual property (IP). The diversity of language, spelling mistakes, colloquial expressions, and even ambiguous keywords can prevent natural language processing (NLP) systems from correctly understanding user needs [2-3]. From a system development perspective, this not only affects the user experience but can also lead to inaccurate information retrieval, ultimately preventing users from obtaining correct answers [4]. Therefore, how to address the impact of these diverse expressions on search results has become a core issue that needs to be solved in system design for the IP field.

Although current large language models (LLMs) and retrieval technologies have made progress in handling user queries, these technologies still face some limitations when applied to the IP field [5-6]. First, LLMs may suffer from hallucination issues when dealing with highly specialized content like patent law and technical terms [7]. Second, existing retrieval models like BM25, while improving retrieval performance through keyword matching and synonym substitution, cannot effectively capture the deeper semantics in user queries [8], our subsequent experimental results also proved this phenomenon. In practical applications, the same question is often expressed by users in multiple different ways (e.g., keyword-based query or concept query) [9-10]. For example, users may ask: "What is the legal status of the invention?" "Patent review of the invention," or "Has the invention been approved?" These queries are semantically similar but essentially the same, but due to the different ways of expression, the vector retrieval model may treat them as different questions instead of the same query [11]. In other words, in the vector space, the embedding model cannot accurately capture the same intent behind them, causing them to be misunderstood as different queries and thus fail to match relevant information.

In addition to the diversity of inquiries, users may also raise intersectional questions that involve multiple fields in a single query [12]. For example, a query not only pertains to the patent review status but also involves legal terms related to patent protection and technical details.

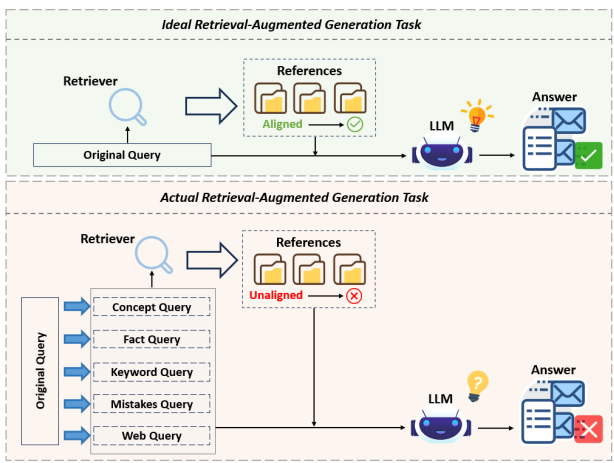

**Fig. 1.** RAG system usage scenarios in the IP field

In the above context, the Retrieval-Augmented Generation (RAG) could serve as an initial solution. For example, when a user asks how to query the status of a confidential patent, the RAG system can not only provide guidance on the review status, but also give relevant patent protection terms, thereby providing the user with a comprehensive answer covering multiple dimensions of information such as legal background, technical implementation and review status. However, the success of RAG is closely tied to the accuracy of the retrieval step. If the retrieval results are inaccurate, the generated answer may contain incorrect legal or technical basis. As shown in Figure 1, the ideal RAG system is designed to retrieve corresponding references. However, various queries often arise in practical use leading to retrieval failures. Thus, how to accurately understand the intention of user queries and ensure that the RAG system in the IP field can correctly respond to users' diverse queries is the primary motivation of this paper. In addition, our secondary motivation is to promote the usability of RAG in the IP field by proposing a simple and generalizable methodology, which facilitates rapid personalized deployment by agencies and solves the practical problem of inaccurate retrieval.

Currently, academia has proposed a variety of methods to optimize the performance of RAG systems. Based on the differences in technical paths, the existing paradigm of method can be grouped into three types: (1) *Data-to-Tune*: This paradigm does not alter the model architecture but focuses on prompt engineering or fine-tuning the model to optimize performance. For example, Kim et al. fine-tuned LLMs by constructing a Chain of Thought (CoT) dataset to explicitly guide the reasoning process, helping models exhibit clearer and more understandable reasoning [13]. (2) *Flow-to-Run*: This paradigm modifies the external retrieval-generation interaction process without changing the model's internal parameters or structure, such as Adaptive-RAG [14]. (3) *Build-to-Learn*: This paradigm modifies the model's internal network architecture or mathematical functions to enhance capabilities, like mixture of experts (MoE) [15]. These paradigms represent different optimization tracks for innovation: *Data-to-Tune* focuses on parameter tuning, *Flow-to-Run* emphasizes process design, and *Build-to-Learn* pursues algorithmic refinement. It is important to note that innovation is not limited to mathematical breakthroughs—choosing the appropriate technical path to meet the demands of different domains is crucial. For example, when aiming for answer diversity in general domains, *Build-to-Learn* might be preferred; however, in some niche fields, where rapid personalized deployment and low-cost fine-tuning are vital for small and medium agencies, *Data-to-Tune* may be the preferred choice.

In this paper, we propose a Multi-Angle Question Generation and Retrieval Fine-Tuning Method (MQG-RFM) using *Data-to-Tune* paradigm. Specifically, we use LLMs as agents to simulate different users to generate inquiries with different preferences. Then, we use hard negative mining to utilize generated inquiries as labels to fine-tune the retrieval model to improve the retrieval ability of the RAG system. Notably, we do not rely on complex mathematical constructions or reshaping the network architecture. Instead, we combine prompt engineering with fine-tuning, providing a straightforward yet effective method to solve difficult problems in the Q&A scenario of the IP field. Through our method, the retrieval model learns how to map different questions to the same answer in the vector space, enhancing its adaptability to diverse queries. Experimental results show that MQG-RFM significantly improves retrieval accuracy and answer generation quality on the Taiwan patent Q&A dataset. This paper makes the following contributions:

(1) A novel methodology MQG-RFM is proposed to utilize LLMs to simulate diverse user questions to generate data for fine-tuning, solving the problem of existing RAG systems in IP filed being unable to handle multiple question expressions.

(2) A simple and generalizable solution is provided without complex architectural changes, our approach offers a practical, low-cost, and highly efficient solution that can be rapidly adapted to various real-world scenarios in the IP field.

(3) Improves the retrieval model's ability to map semantically similar queries expressed in various ways, thereby enhancing the effectiveness of RAG systems in matching user queries with relevant patent information.

## 2. Related Work

According to the hierarchical analysis theory of model optimization in the field of machine learning [16], the improvement of model performance can be achieved by adjusting parameters, optimizing processes or reconstructing architectures. Based on the hierarchical analysis theory, from a perspective of technical paths, the optimization of RAG can also be summarized into these three paradigms: *Data-to-Tune*, *Flow-to-Run*, and *Build-to-Learn*. These paradigms improve RAG performance from the perspectives of parameter tuning, process design, and architectural transformation. The following will introduce the definition, related works, and pros and cons of each paradigm, and compare their essential differences and industrial adaptability, finally reveal the suitable paradigm in the IP field.

### 2.1. Data-to-Tune

The paradigm of *Data-to-Tune* emphasizes optimizing models through data augmentation, hint engineering, or parameter fine-tuning without changing the model's internal architecture or external interaction process. The *Data-to-Tune* approach benefits from its simplicity and flexibility as it does not require structural changes to the model. For example, Yu et al. train the augmentation-adapted retriever (AAR) using preference learning on the source language model to better adapt to a target model (e.g., migrating from Flan-T5 to InstructGPT) to reduce the adaptation cost of heterogeneous models [17]. Similarly, Mao et al. proposed a query rewriting method named RaFe based on reranker feedback to optimize RAG [18]. By first fine-tuning the model to generate rewritten queries and then using reranker scores for feedback training, their fine-tuning method and prompt engineering enhances the model's ability to generate queries better aligned with retrieval targets [18]. RaFe does not require changes to the model architecture or external processes, and only optimizes the model through data enhancement and feedback training, which demonstrates the effectiveness of the *Data-to-Tune* paradigm for RAG system enhancement. Another example is Self-Knowledge guided Retrieval augmentation (SKR) proposed by Wang et al. that trains a mechanism to adaptively decide whether to use the retriever and identify the most similar queries from the training data based on the input query, thereby reducing the interference of irrelevant retrieval on the generated results to optimize RAG [19].

### 2.2. Flow-to-Run

*Flow-to-Run* focuses on the workflow design and inter-module collaboration strategies to optimize RAG effects. Rather than modifying the network architecture of models, this paradigm enhances performance by improving how modules interact. As Jeong et al. emphasized, the generation module and the retrieval module do not operate independently, but are collaborative work [14]. Jeong et al. believe that a good RAG workflow is that generation module can dynamically decide whether to call the retrieval module based on task requirements, while the retrieval module provides customized information based on the requirements of the generation model to optimize the generation results [14]. Therefore, Jeong et al. proposed Adaptive-RAG, which optimizes the RAG system by dynamically selecting retrieval strategies and collaborating with the generation module through a classifier [14]. In the specific application, Louis et al. proposed the Retrieve-Then-Read (R2R) process to apply the RAG to the legal field [20]. They effectively verified the feasibility of RAG in vertical fields. Another noteworthy work is the summarized retrieval (SuRe) framework by Kim et al., which enhances module synergy through a structured process that includes candidate answer generation, document summarization, and answer verification [21]. Specifically, the SuRe framework enables each module to play a specific role in the workflow through the process design of candidate answer generation, retrieval document summary, and answer verification.

### 2.3. Build-to-Learn

The *Build-to-Learn* paradigm refers to modifying the neural network architecture or related functions within the model to make the model more capable of learning. The *Build-to-Learn* paradigm can significantly enhance model capabilities, but it often requires complex architectural design and massive computational resources for training and validating. OPEN-RAG is a representative work of the *Build-to-Learn* paradigm, which converts LLMs into a parameter-efficient sparse MoE and enhances the generation ability by introducing special reflection tokens [22]. Similarly, Asai et al. proposed a method named Self-RAG that introduced reflection tokens inside the model to modify the generation process, allowing the model to perform self-evaluation at each generation step and adjust the generation strategy based on the evaluation results [23]. In addition, the REPLUG method by Shi et al. innovates the retriever's training process by calculating the similarity between each retrieved document and the query during the training phase, and then uses the Softmax method to calculate the selection probability of K documents [24]. REPLUG has made structural innovations for the retriever, enabling it to select the best document.

TABLE I
Comparison of Paradigms and Their Suitability for IP

| Aspect | Data-to-Tune | Flow-to-Run | Build-to-Learn |
|---|---|---|---|
| Object | Parameter/Data | Process/Workflow | Architecture/Function |
| Cost | Low (Data-centric) | Moderate (Workflow design) | High (Structural redesign) |
| Adaptability | Fast domain adaptation | Dynamic query handling | High-precision tasks |
| Limitation | Data coverage dependency | Latency-accuracy tradeoff | Deployment scalability |

### 2.4. Comparison and Suitability for IP

Each of the three paradigms offers unique advantages and limitations. The paradigm of *Data-to-Tune* is ideal for scenarios where model architecture does not require changes, and prompt engineering or fine-tuning with high-quality data is feasible. *Flow to Run* paradigm can design workflow for LLM interactions based on the specific needs of the scenario. Finally, the *Build-to-Learn* paradigm provides innovation in neural network architecture but at the cost of increased complexity and resource requirements to train and validate the feasibility of the reconstructed network.

In the context of IP, several core requirements of NLP applications have been identified through extensive research based on practical considerations [25-27], including (1) Handling expression diversity: Most user queries contain spelling errors or colloquial expressions; (2) Traceability of legal basis: The answer must be strictly related to the provisions of the Patent Law; (3) Feasibility of small and medium-sized institutions: Efficient deployment needs to be achieved with limited computing power. These requirements are illustrated in Table I, which compares the three paradigms in terms of their adaptability to the IP field. Similarly, Aristodemou et al. also pointed out that NLP systems in the IP field require processing complex legal language and terminology, which places high demands on the adaptability and accuracy of models [26]. In addition, Ilin et al. explored the need to consider computational resource limitations and efficient deployment when deploying LLMs in small enterprises [27].

Therefore, it is crucial to choose the appropriate paradigm for different scenarios. Among the three paradigms, *Data to Tune* is the optimal choice due to its low cost, high flexibility, and fast iteration capability. Although there are many existing methods for the *Data to Tune* paradigm, such as AAR and RaFe, their strong dependence on data coverage has not been effectively addressed in the field of IP. Existing methods often cannot cover all relevant term variants, resulting in inaccurate results and missing information in RAG system queries. Therefore, we propose the MQG-RFM method in this paper, which inherits the efficiency advantages of the *Data to Tune* paradigm and fine-tunes the retrieval model by simulating multidimensional problems to solve these gaps.

## 3. Proposed Method

The methodology of MQG-RFM involves leveraging LLMs as agents simulating various users in IP field for query generation, hard negative mining for data augmentation, and fine-tuning retrieval model, as illustrated in figure 2.

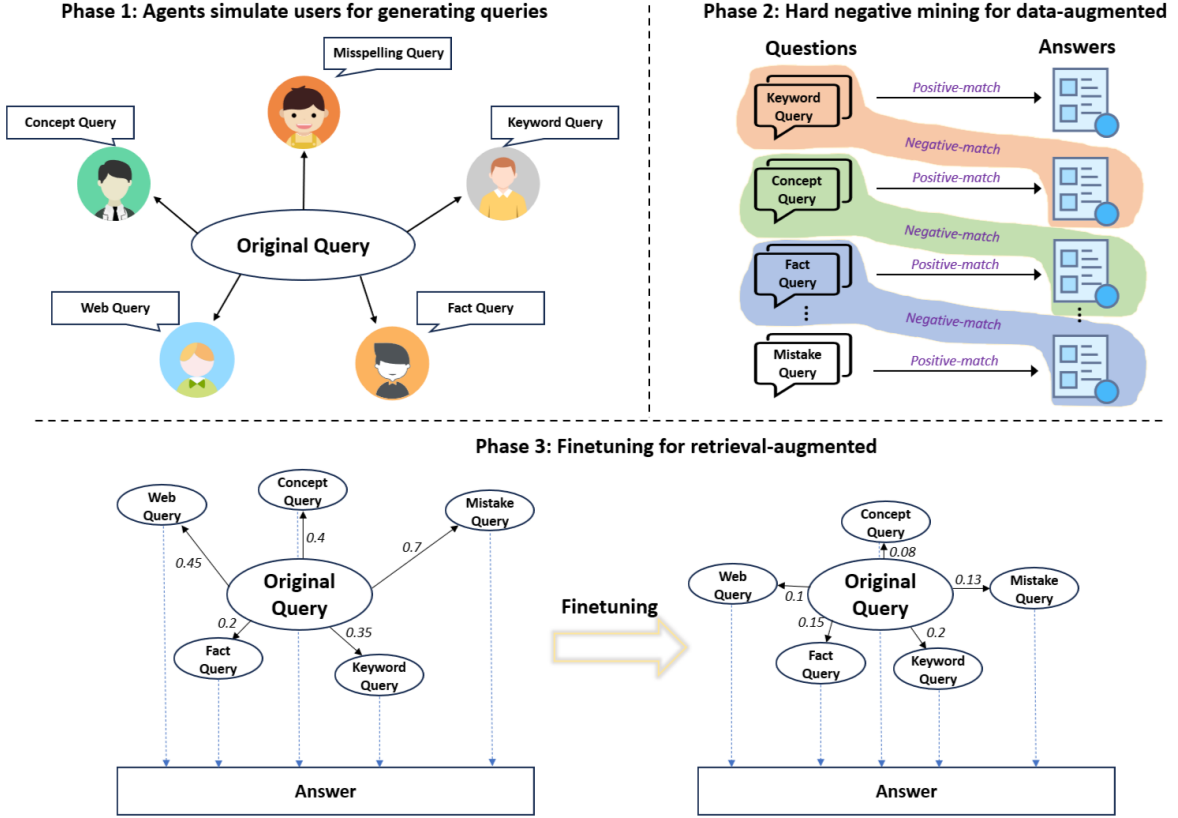

**Fig. 2.** Implementation Process of MQG-RFM

### 3.1. Agent simulation

Our approach begins by utilizing a LLM as an agent to generate multiple queries based on the original query $q_{\text{orig}}$ in the logs, which goal is to simulate different user preferences by generating various types of queries. Specifically, for each original query $q_0$, the agent generates a set of $k$ queries $Q_{\text{gen}} = \{q^1_{\text{type}_i}, q^2_{\text{type}_i}, \ldots, q^k_{\text{type}_i}\}$ for query types with different preferences $T_i$ separately. We define the set of query types $T = \{T_1, T_2, \ldots, T_k\}$, where each type corresponds to a different seeking behavior. Each query type $T_i$ is used to generate queries $q_{\text{gen}_i}$ based on the original query $q_0$:

$$q^{T_i}_{\text{gen}_j} = \text{LLM}(q_0, T_i) \qquad (1)$$

where $q^{T_i}_{\text{gen}_j}$ represents the $i$-th generated query of type $T$. By generating $k$ queries per query type $T_i$, we ensure that each user intent is wellrepresented.

### 3.2. Data Augmentation

After generating the diverse queries, we proceed to the data augmentation phase using hard negative mining. This process involves pairing each generated query $q^{T_i}_{\text{gen}_j}$ with both positive and negative answers to create augmented training examples.

**Positive Example**: The answer $a_0$ corresponding to the original query $q_0$ is used as the positive example for each generated query, which is assumed to be correct and relevant for the generated query:

$$\text{Pos}^{T_i}_{\text{gen}_j} = (q_0, a_0) \text{ for each } q^{T_i}_{\text{gen}_j} \qquad (2)$$

**Negative Example**: For each generated query $q^{T_i}_{\text{gen}_j}$, we create a negative example by selecting an answer $a_{\text{neg}}$ that do not correspond to $q^{T_i}_{\text{gen}_j}$. To ensure the difficulty, we randomly select $a_{\text{neg}}$ from the set of answers that are not the correct answer for $q^{T_i}_{\text{gen}_j}$:

$$Neg^{T_i}_{\text{gen}_j} = (q_0, a_{\text{neg}}) \text{ for each } q^{T_i}_{\text{gen}_j} \qquad (3)$$

By generating positive and negative examples, we get an augmented dataset $(\text{Pos}^{T_i}_{\text{gen}_j}, \text{Neg}^{T_i}_{\text{gen}_j})$ that help the model distinguish between correct and incorrect answers.

### 3.3. Finetuning

The final stage in MQG-RFM involves using the augmented dataset $(\text{Pos}^{T_i}_{\text{gin}_j}, Neg^{T_i}_{\text{gen}_j})$ to fine-tune the retriever. The positive and negative examples generated from the previous step are incorporated into the model's training set. The goal of this phase is to fine-tune the retrieval model $R(\theta)$, parameterized by $\theta$, to correctly retrieve answers. The loss function $\mathcal{L}(\theta)$ is defined for fine-tuning as:

$$\mathcal{L}(\theta) = -\frac{1}{n}\sum_{j=1}^{n} \log \frac{\exp\left(\text{sim}\left(q_{\text{gen}_j}^{T_i}, \text{Pos}_{\text{gen}_j}^{T_i}\right)\right)}{\sum_{j=1}^{B}\exp\Delta\left(\text{sim}\left(q_{\text{gen}_j}^{T_i}, \text{Neg}_{\text{gen}_j}^{T_i}\right)\right)} \quad (4)$$

where $B$ is the batch size. The loss function encourages the model to maximize the probability of retrieving the correct answer and minimize the probability of retrieving the incorrect answer. This leads to a model that is better able to distinguish between relevant and irrelevant answers, improving the quality of the retrieval process for RAG system.

## 4. Data

### 4.1. Dataset

We use the real dataset related to patents and IP provided by the Taiwan government:

**Patent Consultation (PC) Q&A**: This dataset is from the frequent Q&A of the Taiwan Patent Service Center. The content includes basic knowledge of patents, patent procedures, formal examination, change of application, annual fees, changes in patent rights, correction of patent rights, patent retrieval, patent attorney management and other Q&A related to IP business.

**Novel Patent Technology Report (NPTR) Q&A**: This dataset contains frequent Q&A about novel patent technology reports in Taiwan, including the legal basis, acceptance and comparison of novel patent technology reports, and relevant regulations.

### 4.2. Data Analysis

We developed a platform to facilitate data management for institutions shown in Appendix A, enabling them to efficiently manage logs and perform analysis. The platform allows administrators to leverage LLM to play different user roles, thereby generating a range of queries for analysis based on specific user preferences.

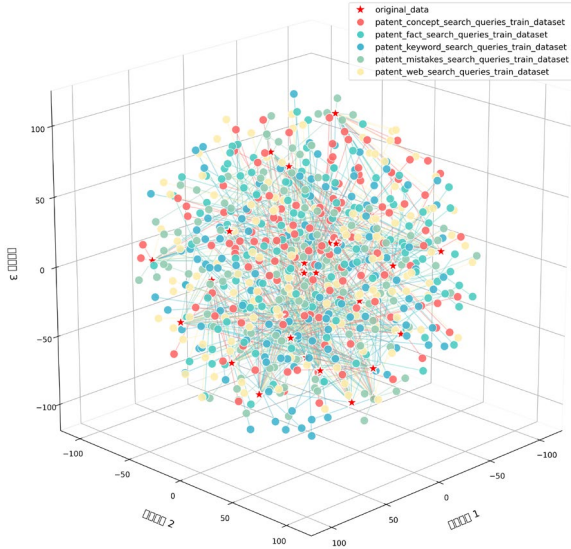

**Fig. 3.** Spatial distance between the generated query and the original query

In our study, we employed GPT-4 as an agent to simulate different user query preferences. To demonstrate the diversity of the generated queries compared to the original ones, we visualized the semantic distance between the generated and original queries using t-SNE. As shown in figure 4, the t-SNE plot maps these queries to a vector space, with each query positioned according to its semantic meaning. From the plot, it is evident that each generated query maintains a certain distance from the original queries, confirming that the model successfully produces diverse queries that differ from the initial set.

## 5. Experimental Setup

For MQG-RFM, we employed the Dmeta-embedding-zh [1] after finetuning with 5 epoch as the retriever. The max input length of generator model is set to 4096. For approaches not utilizing custom-defined prompts, we applied a connected prompt, which is shown in the Appendix B. The methods in the baselines adopt the same settings and hyperparameters used by Jin et al. in their work [28]. All experiments are carried out on 4 NVIDIA 4090 GPUs.

### 5.1. Metrics

We evaluate the performance of RAG methods using the following quantitative metrics:

**ROUGE-1**: It measures the degree of overlap between the generated text and the 1-gram in the reference text, emphasizing word-level coverage [29]. Its formula is:

$$\text{ROUGE-1} = \frac{\sum_{w \in \text{Ref}} \text{Count}_{1\text{-gram}}(w)}{\sum_{w \in \text{Ref}} \text{Count}(w)} \quad (5)$$

where $\text{Count}_{1\text{-gram}}(w)$ is the number of 1-gram matches between the generated text and the reference text and $\text{Count}(w)$ is the total number of 1-grams in the reference text.

**ROUGE-L**: This indicator is based on the longest common subsequence (LCS) to calculate F-measure, which reflects the similarity in semantic order and structure between the generated text and the reference text [29]. Its formula is:

$$R_{LCS} = \frac{LCS(X,Y)}{m} \quad (6)$$

$$P_{LCS} = \frac{LCS(X,Y)}{n} \quad (7)$$

$$F_{LCS} = \frac{(1+\beta^2)R_{LCS}P_{LCS}}{\beta^2 R_{LCS} + P_{LCS}} \quad (8)$$

where $LCS(X,Y)$ is the longest common subsequence length for generating text $X$ and reference text $Y$, $m$ is the length of reference text, $n$ is the length of generated text, and $\beta$ is the ratio of precision to recall.

**BLEU:** BLEU mainly focuses on the n-gram accuracy of generated text, and also reflects semantics and fluency [30]. The following is the formulation of this metric:

$$\text{BLEU} = \text{BP} \cdot \exp\left(\sum_{n=1}^{N} w_n \cdot \log p_n\right), \text{BP} = \begin{cases} 1 & \text{if } c > r \\ e^{(1-\frac{r}{c})} & \text{if } c \leq r \end{cases} \quad (9)$$

where $w_n$ is the weight of n-grams, $m$ is the length of reference text, $BP$ is the length penalty factor, $c$ is the length of the generated text, and $r$ is the length of the reference text.

**Hit Rate**: The hit rate is used to measure the proportion of queries that find the correct answer in the Top-K retrieved documents [31]. It effectively quantifies the success rate of the system in providing the correct answer in the initial set of guesses. The rate is calculated using the formula:

---

[1] https://huggingface.co/DMetaSoul/Dmeta-embedding-zh

$$\text{Hit Rate} = \frac{\text{Number of hits in top-k}}{\text{Total queries}} \quad (10)$$

**BERT-P:** It measures how much of the predicted text matches the reference in terms of contextual embeddings. A high value indicates that the predicted text closely aligns with the reference text. It can be calculated using the formula:

$$\text{BERT-P} = \frac{1}{|C|} \sum_{c \in C} \max_{r \in R} \text{sim}(c, r) \quad (11)$$

where $C$ is the set of tokens in the predicted text, $R$ is the set of tokens in the reference text, and $\text{sim}(c, r)$ is the contextual embedding similarity between $c$ and $r$.

**BERT-R:** It measures how much of the reference text is captured by the predicted text. A high value indicates that the reference text is well covered by the prediction. It can be calculated using the formula:

$$\text{BERT-R} = \frac{1}{|R|} \sum_{r \in R} \max_{c \in C} \text{sim}(r, c) \quad (12)$$

where all notations are as above, but now each token $r$ in the reference looks for its most similar token $c$ in the predicted text.

**BERT-F1:** It is the harmonic mean of BERT-P and BERT-R, which balances both metrics. A high value indicates that both coverage (BERT-P) and completeness (BERT-R) are simultaneously satisfied. It can be calculated using the formula:

$$\text{BERT-F1} = 2 \cdot \frac{\text{BERT-P} \times \text{BERT-R}}{\text{BERT-P} + \text{BERT-R}} \quad (13)$$

**NDCG:** It measures the ranking quality of retrieved documents by considering both the relevance of the documents and their position in the ranked list [32]. A high value indicates that highly relevant documents appear at the top of the ranking more often. It can be calculated using the formula:

$$\text{DCG@}k = \sum_{i=1}^{k} \frac{2^{\text{rel}_i} - 1}{\log_2(i + 1)} \quad (14)$$

$$\text{IDCG@}k = \sum_{i=1}^{k} \frac{2^{\text{rel}_i^{(\text{ideal})}} - 1}{\log_2(i + 1)} \quad (15)$$

$$\text{NDCG@}k = \frac{\text{DCG@}k}{\text{IDCG@}k} \quad (16)$$

$\text{rel}_i$ is the relevance grade of the $i$-th retrieved document, and $\text{rel}_i^{(\text{ideal})}$ is the relevance grade in the ideal ranking.

**Precision:** It measures the proportion of relevant documents among the retrieved documents. A high value indicates that the system returns mostly relevant documents. It can be calculated using the formula:

$$\text{Precision} = \frac{\text{TP}}{\text{TP} + \text{FP}} \quad (17)$$

where TP (True Positives) is the number of retrieved relevant documents, and FP (False Positives) is the number of retrieved irrelevant documents.

### 5.2. Baselines

In our work, we compare our proposed MQG-RFM with several state-of-the-art methods that adopt the *Data-to-Tune*, *Flow-to-Run*, and *Build-to-Learn* paradigms. These methods include: AAR [17], SKR [19], Self-RAG [23], Adaptive-RAG [14], and SuRe [21].

## 6. Evaluation

### 6.1. Retrieval Model Comparison

We compare the retrieval performance of various embedding models bge-large-zh-v1.5 [2], Dmeta-embedding-zh, stella-base-zh-v3-1792d [3], PatentSBERTa [4] on PC and NPTR datasets, as shown in Table II. Notably, even though PatentSBERTa is specialized for patent text, its performance remains lower than expected. In contrast, general-purpose embeddings like bge-large-zh-v1.5 and Dmeta-embedding-zh exhibit greater in certain cases, but they still fall short of accurately retrieving patent documentation. Overall, the results demonstrate that our method achieves substantially higher scores across most metrics.

TABLE II
Comparison of the retrieval performance of different retrievers

| Data | Model | MRR | Precision@3 | Hit@1 | Hit@3 | NDCG@1 | NDCG@3 |
|---|---|---|---|---|---|---|---|
| PC | bge-large-zh | 0.673 | 0.250 | 0.563 | 0.751 | 0.563 | 0.675 |
| | stella-base-zh | 0.670 | 0.244 | 0.571 | 0.734 | 0.571 | 0.668 |
| | Dmeta-embedding | 0.712 | 0.259 | 0.621 | 0.779 | 0.621 | 0.713 |
| | PatentSBERTa | 0.218 | 0.084 | 0.148 | 0.252 | 0.148 | 0.807 |
| | **MQG-RFM** | **0.749** | **0.288** | **0.663** | **0.857** | **0.663** | **0.777** |
| NPTR | bge-large-zh | 0.685 | 0.256 | 0.576 | 0.769 | 0.576 | 0.693 |
| | stella-base-zh | 0.675 | 0.282 | 0.538 | 0.846 | 0.538 | 0.707 |
| | Dmeta-embedding | 0.722 | 0.269 | 0.615 | 0.807 | 0.615 | 0.726 |
| | PatentSBERTa | 0.429 | 0.141 | 0.307 | 0.423 | 0.307 | 0.380 |
| | **MQG-RFM** | **0.980** | **0.333** | **0.961** | **1** | **0.961** | **0.985** |

### 6.2. Method Comparison

In order to compare the advancement of the proposed method in RAG, we also compared it with various state-of-the-art methods in terms of retrieval and generation. Table III compares retrieval performance across various RAG methods on the PC and NPTR datasets, while Table IV focuses on the impact of these methods on downstream generation task using DeepSeek[5]. From Table III, we observe that traditional approaches like AAR and SKR yield relatively low Hit@1 and MRR values, suggesting that they often fail to return highly relevant documents at the top of the rank. Methods such as SuRe and Self-RAG improve retrieval precision to some extent, but still low. Adaptive-RAG exhibits better robustness by refining the interplay between retrieval and generation in an iterative fashion, yet its Hit@1 remains below 0.25 on the PC dataset, indicating an ongoing difficulty in quickly pinpointing the most relevant documents. In contrast, MQG-RFM achieves substantial improvements across all metrics: on the PC dataset, Hit@1 improves from 0.249 (Self-RAG) to 0.663, while MRR surges from below 0.185 to 0.749. Even more striking are the NPTR results, where our method attains near-perfect retrieval accuracy with Hit@1 reaching 0.961 and Hit@3 achieving a perfect 1.000, underscoring the model's capacity to handle complex patent queries that interleave technical and legal terminology.

---

[2] https://huggingface.co/BAAI/bge-large-zh-v1.5

[3] https://huggingface.co/infgrad/stella-base-zh-v3-1792d

[4] https://huggingface.co/AI-Growth-Lab/PatentSBERTa

[5] https://huggingface.co/deepseek-ai/DeepSeek-R1-Distill-Qwen-14B

TABLE III
Comparison of the retrieval performance of different RAG methods

| Data | Method | MRR | Precision@3 | Hit@1 | Hit@3 | NDCG@1 | NDCG@3 |
|---|---|---|---|---|---|---|---|
| PC | AAR [17] | 0.089 | 0.082 | 0.184 | 0.038 | 0.036 | 0.022 |
| | SKR [19] | 0.089 | 0.082 | 0.173 | 0.038 | 0.036 | 0.022 |
| | SuRe [21] | 0.400 | 0.168 | 0.320 | 0.505 | 0.320 | 0.467 |
| | Self-RAG [23] | 0.185 | 0.108 | 0.249 | 0.113 | 0.076 | 0.061 |
| | Adaptive-RAG [14] | 0.128 | 0.121 | 0.232 | 0.082 | 0.071 | 0.049 |
| | **MQG-RFM** | **0.749** | **0.288** | **0.663** | **0.857** | **0.663** | **0.777** |
| NPTR | AAR [17] | 0.150 | 0.166 | 0.265 | 0.098 | 0.093 | 0.064 |
| | SKR [19] | 0.141 | 0.157 | 0.206 | 0.056 | 0.082 | 0.055 |
| | SuRe [21] | 0.224 | 0.089 | 0.192 | 0.269 | 0.192 | 0.235 |
| | Self-RAG [23] | 0.152 | 0.046 | 0.225 | 0.099 | 0.035 | 0.028 |
| | Adaptive-RAG [14] | 0.154 | 0.165 | 0.235 | 0.078 | 0.092 | 0.063 |
| | **MQG-RFM** | **0.980** | **0.333** | **0.961** | **1** | **0.961** | **0.985** |

These superior retrieval results carry over to downstream generation performance, as shown in Table IV. Methods with weaker retrieval components tend to underperform in both surface-level (ROUGE, BLEU) and semantic-level (BERT-based) metrics, primarily because incorrect or incomplete references feed into the generation module. Among the baselines, Self-RAG and Adaptive-RAG produce moderate improvements in ROUGE and BLEU scores on the PC dataset but still do not fully close the gap to MQG-RFM. By contrast, MQG-RFM not only raises ROUGE-1 from below 0.25 to above 0.26 on PC but also substantially boosts ROUGE-2 and BLEU-1 through BLEU-4 scores, reflecting better lexical and n-gram alignment with the ground truth. More importantly, the BERT-based scores (BERT-P, BERT-R, BERT-F1) also exhibit the largest gains under MQG-RFM, indicating that our model retrieves references whose content is genuinely closer in meaning to the target answers. On the NPTR dataset, where patent queries can be more specialized, MQG-RFM's advantage further widens, raising ROUGE and BLEU metrics by clear margins over baselines.

Overall, the findings validate that stronger retrieval is a linchpin for improved RAG performance, as higher Hit@1 and MRR directly translate into more contextually aligned knowledge for generation. By leveraging simulated multi-angle queries to fine-tune the retriever, MQG-RFM successfully maps diverse user expressions to the same underlying intent, thus ensuring that patent-specific topics are accurately surfaced and integrated into the generated answers.

TABLE IV
Comparison of the performance of different RAG methods in generation using DeepSeek

| Data | Method | ROUGE-1 | ROUGE-L | BLEU-1 | BLEU-2 | BLEU-3 | BLEU-4 | BERT-P | BERT-R | BERT-F1 |
|---|---|---|---|---|---|---|---|---|---|---|
| PC | AAR [17] | 0.184 | 0.089 | 0.082 | 0.036 | 0.022 | 0.014 | 0.578 | 0.655 | 0.612 |
| | SKR [19] | 0.173 | 0.089 | 0.082 | 0.036 | 0.022 | 0.014 | 0.578 | 0.655 | 0.612 |
| | SuRe [21] | 0.033 | 0.026 | 0.007 | 0.006 | 0.006 | 0.005 | 0.501 | 0.357 | 0.413 |
| | Self-RAG [23] | 0.249 | **0.185** | 0.108 | 0.076 | 0.061 | 0.050 | 0.701 | 0.629 | 0.661 |
| | Adaptive-RAG [14] | 0.232 | 0.128 | 0.121 | 0.071 | 0.049 | 0.035 | 0.617 | 0.704 | 0.655 |
| | **MQG-RFM** | **0.265** | 0.140 | **0.132** | **0.084** | **0.061** | **0.052** | 0.629 | **0.725** | **0.671** |
| NPTR | AAR [17] | 0.265 | 0.150 | 0.166 | 0.093 | 0.064 | 0.044 | 0.646 | 0.682 | 0.661 |
| | SKR [19] | 0.206 | 0.141 | 0.157 | 0.082 | 0.055 | 0.037 | 0.639 | 0.664 | 0.649 |
| | SuRe [21] | 0.026 | 0.020 | 0.005 | 0.004 | 0.003 | 0.002 | 0.499 | 0.342 | 0.404 |
| | Self-RAG [23] | 0.225 | 0.152 | 0.046 | 0.035 | 0.028 | 0.022 | 0.720 | 0.606 | 0.657 |
| | Adaptive-RAG [14] | 0.235 | 0.154 | 0.165 | 0.092 | 0.063 | 0.045 | 0.661 | 0.679 | 0.668 |
| | **MQG-RFM** | **0.406** | **0.219** | **0.235** | **0.159** | **0.124** | **0.098** | **0.723** | **0.724** | **0.715** |

6.3. Ablation Study

To investigate the effectiveness of our strategy in MQG-RFM, we compare method without fine-tuning on both generation and retrieval metrics, as shown in figure 4 and 5.

On the retrieval side (Figure 4), removing the fine-tuning stage results in a marked drop in all metrics. Specifically, Hit@1 plummets from 0.962 to 0.192, and MRR from 0.981 to 0.224, indicating that without fine-tuning, the model struggles to identify the most relevant documents within the top ranks. A similar pattern emerges in P@3 and NDCG scores, where the decline is similarly pronounced. These results confirm that the multi-angle question generation process, followed by fine-tuning, is crucial for aligning diverse user queries with the correct patent documents in the vector space.

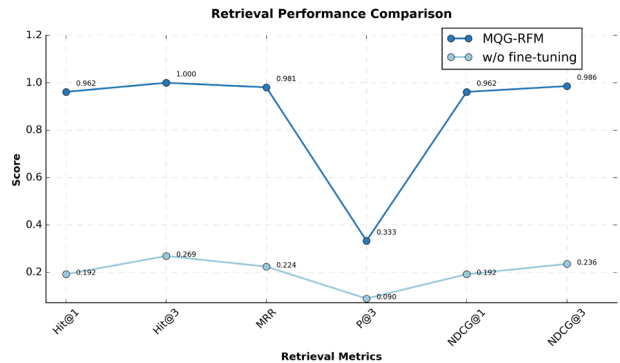

**Fig. 4.** Ablation Experiment about Retrieval

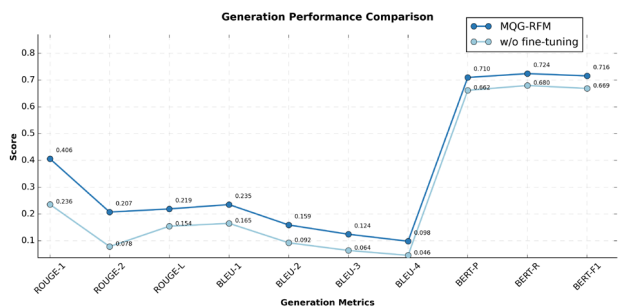

**Fig. 5.** Ablation Experiment about Generation

A comparable trend can be observed in Figure 5 for downstream answer generation. Without fine-tuning, both n-gram-based metrics (ROUGE and BLEU) and semantic metrics (BERT-based precision, recall, and F1) exhibit substantially lower values. For instance, ROUGE-1 decreases from 0.406 to 0.236, while BLEU-1 drops from 0.235 to 0.165. The gap is similarly pronounced in BERT-based scores, where the F1 metric diminishes from 0.716 to 0.669. These discrepancies underscore the importance of accurate retrieval in enabling the generator to access reliable contextual information. When the retriever is not fine-tuned, it often supplies suboptimal references, leading to less coherent and semantically weaker generated answers.

Overall, this ablation study highlights the critical role of fine-tuning in the MQG-RFM pipeline. Our proposed multi-angle question generation and retrieval fine-tuning scheme ensures robust performance across both retrieval and generation metrics, reinforcing its practical utility for real-world patent search and Q&A scenarios.

### 6.4. Robustness Test

To further validate the generalizability of our method, we replaced the original generation component with Qwen-14B[6] and re-evaluated the system under the same RAG framework. The results presented in Table V show that MQG-RFM maintains a clear lead over baseline methods on both the PC and NPTR datasets across all evaluated metrics. For instance, on the NPTR dataset, MQG-RFM raises BERT-F1 from below 0.52 across baselines to 0.676, suggesting that, even when the generation model's underlying architecture changes, the system still benefits from our fine-tuned retriever and remains capable of delivering answers whose core content aligns closely with ground truth. This consistency holds true for other metrics such as BLEU-1 and BLEU-2, which demonstrate that our approach supplies Qwen-14B with more contextually relevant references, enabling higher-quality content coverage in the generated responses.

Collectively, these findings confirm the robustness and portability of MQG-RFM in diverse RAG setups. Even with a different LLM at the generation stage, our method consistently outperforms domain-specific and general-purpose baselines. Such results reinforce the view that centering on multi-angle question generation—coupled with fine-tuning—significantly enhances retrieval and, by extension, improves downstream question-answering tasks regardless of which large language model is ultimately employed.

TABLE V
Comparison of the performance of different RAG methods in generation using Qwen

| Data | Method | ROUGE-1 | ROUGE-L | BLEU-1 | BLEU-2 | BLEU-3 | BLEU-4 | BERT-P | BERT-R | BERT-F1 |
|---|---|---|---|---|---|---|---|---|---|---|
| PC | AAR [17] | 0.094 | 0.055 | 0.010 | 0.006 | 0.005 | 0.003 | **0.611** | **0.445** | **0.513** |
| | SKR [19] | 0.054 | 0.039 | 0.011 | 0.006 | 0.004 | 0.003 | 0.543 | 0.395 | 0.455 |
| | SuRe [21] | 0.025 | 0.015 | 0.002 | 0.002 | 0.001 | 0.001 | 0.481 | 0.315 | 0.380 |
| | Self-RAG [23] | 0.008 | 0.006 | 0.003 | 0.002 | 0.002 | 0.001 | 0.502 | 0.38 | 0.437 |
| | Adaptive-RAG [14] | 0.085 | 0.062 | 0.014 | 0.011 | 0.010 | 0.009 | 0.582 | 0.411 | 0.480 |
| | **MQG-RFM** | **0.112** | **0.084** | **0.019** | **0.016** | **0.015** | **0.013** | 0.609 | 0.430 | 0.502 |
| NPTR | AAR [17] | 0.094 | 0.055 | 0.010 | 0.006 | 0.005 | 0.003 | 0.611 | 0.445 | 0.513 |
| | SKR [19] | 0.072 | 0.044 | 0.001 | 0.002 | 0.002 | 0.001 | 0.632 | 0.450 | 0.524 |
| | SuRe [21] | 0.025 | 0.015 | 0.002 | 0.002 | 0.001 | 0.001 | 0.481 | 0.315 | 0.380 |
| | Self-RAG [23] | 0.008 | 0.004 | 0.003 | 0.002 | 0.001 | 0.001 | 0.493 | 0.395 | 0.437 |
| | Adaptive-RAG [14] | 0.100 | 0.067 | 0.015 | 0.011 | 0.008 | 0.006 | 0.625 | 0.44 | 0.516 |
| | **MQG-RFM** | **0.320** | **0.235** | **0.077** | **0.069** | **0.064** | **0.059** | **0.809** | **0.587** | **0.676** |

---

[6] https://huggingface.co/Qwen/Qwen2.5-14B-Instruct

**6.5. Case Study**

To illustrate how our MQG-RFM approach enhances real-world patent Q&A, we present in Figure 6 a direct comparison between an answer generated by MQG-RFM (Response #1) and an answer produced by a general model lacking our specialized fine-tuning (Response #2). The user's query asks for the differences among invention patents, new patents, and design patents. While both responses offer a broad outline of each patent type, a closer inspection reveals several notable distinctions in depth, accuracy, and domain relevance.

First, MQG-RFM provides a structured breakdown that clearly distinguishes between the three patent categories along multiple dimensions, such as the protected object, technical requirements, and review method. By highlighting key features—such as the technological scope of invention patents, the novelty-oriented improvements in new patents, and the aesthetic focus of design patents. Furthermore, the system articulates these differences with clarity, underscoring how invention patents typically undergo more rigorous technical examination and enjoy broader protection scope, whereas new patents (often "utility models") focus on incremental improvements to form or structure. This structured and context-aware approach aligns well with the actual implementation of patent law in Taiwan, thereby giving users clearer, legally grounded information on what to expect when applying for or managing different patent types within the jurisdiction.

By contrast, Response #2 delivers a broader, more generic explanation that, while not entirely incorrect for a global audience, overlooks key procedural differences in Taiwan. For example, the stated protection period does not match local practice, and the answer's references to certain legal tests or examination steps appear mismatched or overly generalized, suggesting the model may be hallucinating details from other jurisdictions. Such omissions and inaccuracies greatly diminish the utility of Response #2 for users specifically looking to navigate the Taiwanese patent system.

Overall, this case study highlights the distinctive advantage of MQG-RFM in handling queries specific to Taiwanese patents. While a one-size-fits-all global approach (e.g., Response #2) may be sufficient for high-level queries, it can introduce hallucination or contextual errors for specialized user needs. In contrast, MQG-RFM's fine-tuning with multi-angle question generation provides more precise, jurisdiction-aware answers, ultimately supporting legal professionals and patent applicants seeking reliable, Taiwan-centric guidance.

## Response Comparison

Question: The difference between invention patents, new patents, and design patents

### MQG-RFM Response #1 ✓ Correct Answer

**1. Protected object:**

Invention patent: protects the creation of technological ideas that utilize natural laws, with a wide range of subject matter, including substances, objects, methods, biological materials, and their uses.

New type patent: It only protects the creation of the shape, structure, or combination of items, and belongs to the category of technological improvement.

Design patent: protects the creation of all or part of the shape, pattern, color, or their combination of an item, emphasizing visual aesthetics and competitiveness of the product.

**2. Technical:**

Invention patents and new patents are related to technicality, focusing on improvements in functionality, technology, manufacturing, and ease of use.

Design patents are not related to technicality, but focus on visual quality, affinity, and high-value expression.

**3. Review method:**

Invention patents and design patents require substantive examination to obtain patent rights.

New patents only require formal examination, therefore their patent rights are uncertain and unstable.

**Conclusion**

Invention patent: Strong in technicality and applicable to a wide range of technical fields.

New type patent: focusing more on the improvement of the shape or structure of the object in terms of technology.

Design patents: focus on visual effects, unrelated to technicality.

### General Response #2 ✗ Incorrect Answer

**1. Scope of protection**

Invention patent: protects creations with technical characteristics.

New type patent: protects small improvements or new designs with technical characteristics.

Design patent: protects the appearance design of a product.

**2. Review Procedure**

Invention patent: requires comprehensive technical examination, including judgment of novelty, progressiveness, and industrial applicability.

New type patents: The examination mainly focuses on novelty and the technical features requested by the patentee, while progressive examination is relatively simplified.

Design patents: The examination mainly focuses on the uniqueness and aesthetic value of the design.

**3. Protection period**

Invention patent: usually protected for 20 years, calculated from the date of application.

New type patent: usually protected for 10 years, calculated from the date of application.

Design patents: usually protected for 10 to 15 years, depending on national regulations.

**Fig. 6.** Case Study

## 7. Limitations & Future Work

While MQG-RFM demonstrates promising results, several limitations should be acknowledged: The effectiveness of MQG-RFM relies heavily on the quality of the LLM used for generating diverse user inquiries. If the LLM fails to capture the nuances of user queries or generates low-quality synthetic data, the fine-tuning process may be compromised; While the method is cost-effective for small and medium-sized agencies, scaling it to larger datasets or more complex retrieval tasks may require additional computational resources and optimization.

Hence, several directions remain open for future research and practical enhancements. For example, future work could extend MQG-RFM to support multilingual patent searches (e.g., European Patent Office filings in multiple languages) and cross-jurisdictional legal interpretations. In addition, while our current hard negative mining effectively improves the model's ability to differentiate between semantically similar queries, future efforts could investigate more sophisticated strategies—such as dynamic sampling of hard negatives based on user feedback or cluster-level hardness to further refine retrieval accuracy. By pursuing these research directions, we aim to make MQG-RFM not only more accurate and efficient but also capable of adapting to the evolving nature of patent systems around the globe. As RAG-based solutions continue to grow in sophistication, our focus will remain on bridging the gap between diverse user queries and the precise, domain-specific knowledge required in the IP sector.

## 8. Conclusion

In this paper, we addressed the critical challenge of handling diverse user queries in the IP field, where spelling errors, colloquial expressions, and ambiguous keywords often hinder accurate information retrieval. By focusing on the *Data-to-Tune* paradigm, our method enhances the ability of RAG systems to map semantically similar queries expressed in various ways to the same answer, thereby improving retrieval accuracy and answer generation quality. Our experimental results on the Taiwan patent dataset demonstrate that MQG-RFM significantly outperforms existing methods in handling diverse query expressions, providing a robust solution for real-world IP retrieval scenarios. The proposed method not only addresses the limitations of current RAG systems in the IP field but also offers a simple, generalizable, and cost-effective approach that can be rapidly deployed by small and medium-sized agencies. By combining prompt engineering with fine-tuning, MQG-RFM bridges the gap between user intent and system understanding, ultimately enhancing the usability and effectiveness of RAG systems in the IP domain.

## 9. References

[1] Hegel, G. W. F., Miller, A. V., & Findlay, J. N. (1977). Phenomenology of spirit (Vol. 632). Oxford: Oxford University Press.

[2] Joshi, P., Santy, S., Budhiraja, A., Bali, K., & Choudhury, M. (2020, July). The State and Fate of Linguistic Diversity and Inclusion in the NLP World. In Proceedings of the 58th Annual Meeting of the Association for Computational Linguistics (pp. 6282-6293).

[3] Liu, S. H., Liao, H. L., Pi, S. M., & Hu, J. W. (2011). Development of a Patent Retrieval and Analysis Platform–A hybrid approach. Expert systems with applications, 38(6), 7864-7868.

[4] Kelly, D. (2009). Methods for Evaluating Interactive Information Retrieval Systems with Users. Information Retrieval, 3(1-2), 1-224.

[5] Kirchhübel, C., & Brown, G. (2024). Intellectual Property Rights at the Training, Development and Generation Stages of Large Language Models. LREC-COLING 2024, 13.

[6] Shalaby, W., & Zadrozny, W. (2019). Patent retrieval: a literature review. Knowledge and Information Systems, 61, 631-660.

[7] Dahl, M., Magesh, V., Suzgun, M., & Ho, D. E. (2024). Large legal fictions: Profiling legal hallucinations in large language models. Journal of Legal Analysis, 16(1), 64-93.

[8] Zhao, W. X., Liu, J., Ren, R., & Wen, J. R. (2024). Dense text retrieval based on pretrained language models: A survey. ACM Transactions on Information Systems, 42(4), 1-60.

[9] Montecchi, T., Russo, D., & Liu, Y. (2013). Searching in Cooperative Patent Classification: Comparison between keyword and concept-based search. Advanced Engineering Informatics, 27(3), 335-345.

[10] Gavankar, C., Li, Y. F., & Ramakrishnan, G. (2016). Explicit query interpretation and diversification for context-driven concept search across ontologies. In The Semantic Web–ISWC 2016: 15th International Semantic Web Conference, Kobe, Japan, October 17–21, 2016, Proceedings, Part I 15 (pp. 271-288). Springer International Publishing.

[11] Ai, Q., Yang, L., Guo, J., & Croft, W. B. (2016, September). Analysis of the paragraph vector model for information retrieval. In Proceedings of the 2016 ACM international conference on the theory of information retrieval (pp. 133-142).

[12] Whalen, R. (2018). Boundary spanning innovation and the patent system: Interdisciplinary challenges for a specialized examination system. Research Policy, 47(7), 1334-1343.

[13] Kim, S., Joo, S., Kim, D., Jang, J., Ye, S., Shin, J., & Seo, M. (2023, December). The CoT Collection: Improving Zero-shot and Few-shot Learning of Language Models via Chain-of-Thought Fine-Tuning. In Proceedings of the 2023 Conference on Empirical Methods in Natural Language Processing (pp. 12685-12708).

[14] Jeong, S., Baek, J., Cho, S., Hwang, S. J., & Park, J. C. (2024, June). Adaptive-RAG: Learning to Adapt Retrieval-Augmented Large Language Models through Question Complexity. In Proceedings of the 2024 Conference of the North American Chapter of the Association for Computational Linguistics: Human Language Technologies (Volume 1: Long Papers) (pp. 7029-7043).

[15] Artetxe, M., Bhosale, S., Goyal, N., Mihaylov, T., Ott, M., Shleifer, S., ... & Stoyanov, V. (2022, December). Efficient Large Scale Language Modeling with Mixtures of Experts. In Proceedings of the 2022 Conference on Empirical Methods in Natural Language Processing (pp. 11699-11732).

[16] Houlsby, N., Giurgiu, A., Jastrzebski, S., Morrone, B., De Laroussilhe, Q., Gesmundo, A., ... & Gelly, S. (2019, May). Parameter-efficient transfer learning for NLP. In International conference on machine learning (pp. 2790-2799). PMLR.

[17] Yu, Z., Xiong, C., Yu, S., & Liu, Z. (2023, July). Augmentation-Adapted Retriever Improves Generalization of Language Models as Generic Plug-In. In Proceedings of the 61st Annual Meeting of the Association for Computational Linguistics (Volume 1: Long Papers) (pp. 2421-2436).

[18] Mao, S., Jiang, Y., Chen, B., Li, X., Wang, P., Wang, X., ... & Zhang, N. (2024, November). RaFe: Ranking Feedback Improves Query Rewriting for RAG. In Findings of the Association for Computational Linguistics: EMNLP 2024 (pp. 884-901).

[19] Wang, Y., Li, P., Sun, M., & Liu, Y. (2023, December). Self-Knowledge Guided Retrieval Augmentation for Large Language Models. In Findings of the Association for Computational Linguistics: EMNLP 2023 (pp. 10303-10315).

[20] Louis, A., van Dijck, G., & Spanakis, G. (2024, March). Interpretable long-form legal question answering with retrieval-augmented large language models. In Proceedings of the AAAI Conference on Artificial Intelligence (Vol. 38, No. 20, pp. 22266-22275).

[21] Kim, J., Nam, J., Mo, S., Park, J., Lee, S. W., Seo, M., ... & Shin, J. (2024). SURE: SUMMARIZING RETRIEVALS USING ANSWER CANDIDATES FOR OPEN-DOMAIN QA OF LLMS. In 12th International Conference on Learning Representations, ICLR 2024.

[22] Islam, S., Rahman, M. A., Hossain, K. T., Hoque, E., Joty, S., & Parvez, M. R. (2024, November). Open-RAG: Enhanced Retrieval Augmented Reasoning with Open-Source Large Language Models. In Findings of the Association for Computational Linguistics: EMNLP 2024 (pp. 14231-14244).

[23] Asai, A., Wu, Z., Wang, Y., Sil, A., & Hajishirzi, H. (2023, October). Self-rag: Learning to retrieve, generate, and critique

through self-reflection. In The Twelfth International Conference on Learning Representations.

[24] Shi, W., Min, S., Yasunaga, M., Seo, M., James, R., Lewis, M., ... & Yih, W. T. (2024, June). REPLUG: Retrieval-Augmented Black-Box Language Models. In Proceedings of the 2024 Conference of the North American Chapter of the Association for Computational Linguistics: Human Language Technologies (Volume 1: Long Papers) (pp. 8364-8377).

[25] Gossen, T., & Nürnberger, A. (2013). Specifics of information retrieval for young users: A survey. Information Processing & Management, 49(4), 739-756.

[26] Aristodemou, L., & Tietze, F. (2018). The state-of-the-art on Intellectual Property Analytics (IPA): A literature review on artificial intelligence, machine learning and deep learning methods for analysing intellectual property (IP) data. World Patent Information, 55, 37-51.

[27] Ilin, I., & Kelli, A. (2024). Natural Language, Legal Hurdles: Navigating the Complexities in Natural Language Processing Development and Application. Journal of the University of Latvia. Law, 17, 44-67.

[28] Jin, J., Zhu, Y., Yang, X., Zhang, C., & Dou, Z. (2024). Flashrag: A modular toolkit for efficient retrieval-augmented generation research. arXiv preprint arXiv:2405.13576.

[29] Lin, C. Y. (2004, July). Rouge: A package for automatic evaluation of summaries. In Text summarization branches out (pp. 74-81).

[30] Papineni, K., Roukos, S., Ward, T., & Zhu, W. J. (2002, July). Bleu: a method for automatic evaluation of machine translation. In Proceedings of the 40th annual meeting of the Association for Computational Linguistics (pp. 311-318).

[31] Ren, R., Ma, J., & Zheng, Z. (2025). Large language model for interpreting research policy using adaptive two-stage retrieval augmented fine-tuning method. Expert Systems with Applications, 127330.

[32] Wang, Y., Wang, L., Li, Y., He, D., & Liu, T. Y. (2013, June). A theoretical analysis of NDCG type ranking measures. In Conference on learning theory (pp. 25-54). PMLR.

**Appendix A Interface of Data Management System**

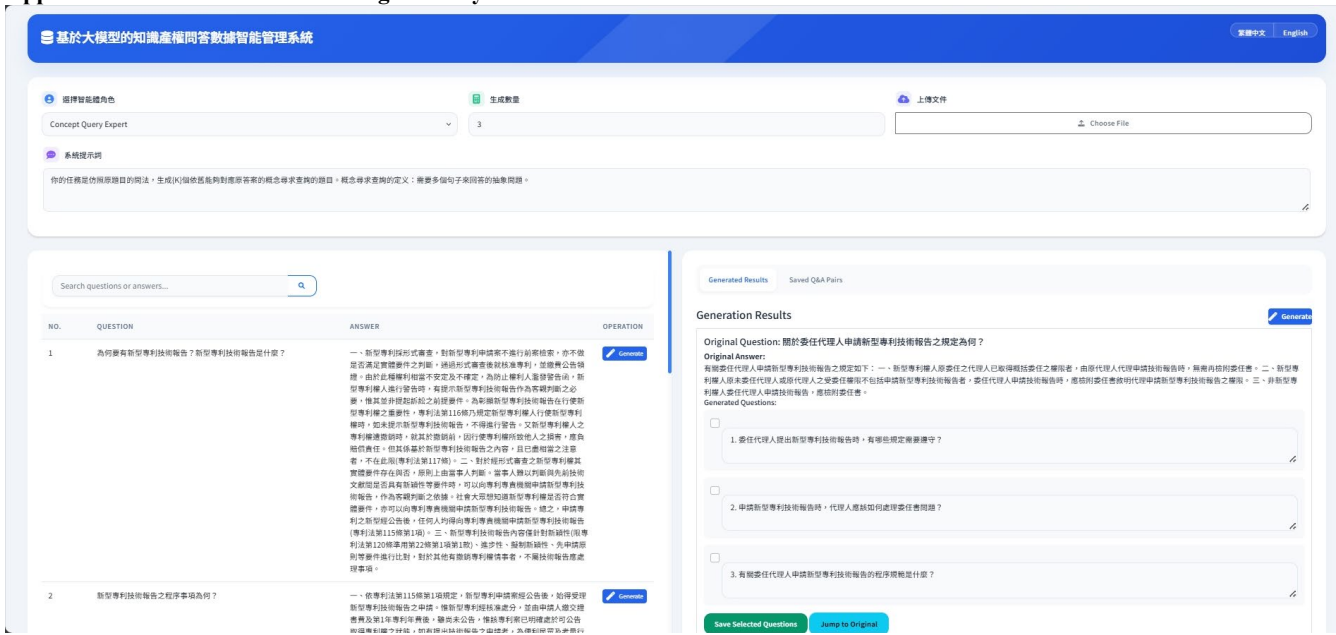

**Appendix B Connected Prompt**

> **Connected Prompt**
>
> **System Prompt:** You are an expert in the field of intellectual property who is good at answering questions based on given documents. Please answer the questions based on the given documents.
>
> **User Prompt:**
> **Question:** [the user query]
> **Context:** [the related context]

**Appendix C Prompts for Generating Queries**

> **Prompt of generating concept seeking query**
>
> 下面是有关台湾便利民众查询及各项申请准备参考使用的专利服务平台平日答询之常见问答:
> ————————
> {context_str}
> ————————
> 你的任务是仿照原题目的问法，生成 {K} 个依旧能够对应原答案的概念寻求查询的题目。概念寻求查询的定义：需要多个句子来回答的抽象问题。
>
> **Translation:**
>
> The following are frequent Q&A about Taiwan's patent service platform, which is used by the public for inquiries and reference in the preparation of various applications:
> ————————
> {context_str}
> ————————
> Your task is to generate {K} concept search queries that still correspond to the original answer, following the method used in the original question. The definition of concept seeking query: an abstract question that requires multiple sentences to answer

**Prompt of generating fact seeking query**

下面是有關台灣便利民眾查詢及各項申請準備參考使用的專利服務平台平日答詢之常見問答：
----------------------
{context_str}
----------------------
你的任務是仿照原題目的問法，生成 {K} 個依舊能夠對應原答案的事實尋求查詢的題目。事實尋求查詢的定義：具有單一、明確答案的查詢。

**Translation:**

The following are frequent Q&A about Taiwan's patent service platform, which is used by the public for inquiries and reference in the preparation of various applications:
----------------------
{context_str}
----------------------
Your task is to generate {K} fact seeking queries that still correspond to the original answer, following the method used in the original question. The definition of fact seeking query: Queries that have a single, unambiguous answer.

**Prompt of generating keyword query**

下面是有關台灣便利民眾查詢及各項申請準備參考使用的專利服務平台平日答詢之常見問答：
----------------------
{context_str}
----------------------
你的任務是仿照原題目的問法，生成 {K} 個依舊能夠對應原答案的關鍵字查詢的題目。關鍵字查詢的定義：僅包含關鍵標識符詞的簡短查詢。

**Translation:**

The following are frequent Q&A about Taiwan's patent service platform, which is used by the public for inquiries and reference in the preparation of various applications:
----------------------
{context_str}
----------------------
Your task is to generate {K} keyword queries that still correspond to the original answer, following the method used in the original question. The definition of keyword query: Short queries containing only key identifier words.

**Prompt of generating query with spelling**

下面是有關台灣便利民眾查詢及各項申請準備參考使用的專利服務平台平日答詢之常見問答：
----------------------
{context_str}
----------------------
你的任務是仿照原題目的問法，生成 {K} 個依舊能夠對應原答案的拼寫錯誤查詢的題目。拼寫錯誤查詢的定義：包含拼寫錯誤、換位和常見拼寫錯誤的查詢。

**Translation:**

The following are frequent Q&A about Taiwan's patent service platform, which is used by the public for inquiries and reference in the preparation of various applications:
----------------------
{context_str}
----------------------
Your task is to generate {K} queries with spelling mistakes that still correspond to the original answer, following the method used in the original question. The definition of query with spelling mistakes: Queries containing misspellings, transpositions, and common spelling errors.

**Prompt of generating web search-like**

下面是有關台灣便利民眾查詢及各項申請準備參考使用的專利服務平台平日答詢之常見問答：
----------------------
{context_str}
----------------------
你的任務是仿照原題目的問法，生成 {K} 個依舊能夠對應原答案的網頁搜索查詢的題目。網頁搜索查詢的定義：類似於通常輸入搜索引擎的簡短查詢轉換。

**Translation:**

The following are frequent Q&A about Taiwan's patent service platform, which is used by the public for inquiries and reference in the preparation of various applications:
----------------------
{context_str}
----------------------
Your task is to generate {K} web search-like queries that still correspond to the original answer, following the method used in the original question. The definition of web search-like query: Similar to short queries commonly entered into search engines.

**Appendix D Examples of Hard Negative Mining**

| classified as *positive* | classified as *negative* |
|---|---|
| **Query**<br>如果我不是專利權人，但我被指控侵犯新型專利，我可以如何申請技術報告來協助處理這一爭議？<br>If I am not the patent owner but I am accused of infringing a utility patent, how can I request a technical report to assist in the dispute?<br>**Document**<br>一、如遇非專利權人有相同或類似商品爲商業上之實施者，爲儘早獲得技術報告，專利權人可以檢附相關証明文件，如專利權人之書麵通知、非專利權人之廣告目錄或其他商業上實施事實之書麵資料，申請新型專利技術報告(參照專利法第 115 條第 5 項及專利法施行細則第 43 條)。本局將於申請後 6 個月內完成新型專利技術報告。\n二、若新型專利技術報告之申請人爲非專利權人，爲協助當事人間侵權爭議之處理，在非專利權人提具事涉專利侵權爭議之相關証明文件，如已遭新型專利權人提出專利侵權之存証信函、涉及專利侵權訴訟案件之起訴書或訴訟傳票等文件資料者，本局亦將優先作成新型專利技術報告。\n三、對於不符合商業上實施規定的技術報告申請案，本局將於受理通知函中敘明"有關商業上實施的主張不符合專利法第 115 條第 5 項之規定"。<br>1. If a non-patent holder has the same or similar products that are commercially implemented, in order to obtain a technical report as soon as possible, the patent holder may attach relevant supporting documents, such as a written notice from the patent holder, an advertising catalog of the non-patent holder, or other written materials of commercial implementation facts, to apply for a new patent technical report (refer to Article 115, Paragraph 5 of the Patent Law and Article 43 of the Patent Law Enforcement Rules). The Office will complete the new patent technical report within 6 months after the application. \n2. If the applicant for the new patent technical report is a non-patent holder, in order to assist in the handling of infringement disputes between the parties, the Office will also give priority to preparing a new patent technical report if the non-patent holder provides relevant supporting documents related to the patent infringement dispute, such as a registered letter of patent infringement filed by the new patent holder, an indictment or a litigation summons involving a patent infringement lawsuit, and other documents and materials. \n3. For technical report applications that do not meet the requirements for commercial implementation, the Office will state in the acceptance notice that "the claim regarding commercial implementation does not meet the requirements of Article 115, Paragraph 5 of the Patent Law." | **Query**<br>如果我不是專利權人，但我被指控侵犯新型專利，我可以如何申請技術報告來協助處理這一爭議？<br>If I am not the patent owner but I am accused of infringing a utility patent, how can I request a technical report to assist in the dispute?<br>**Document**<br>新型專利技術報告申請專利范圍中每一請求項逐項比對后，引用文獻之記載分別說明如下：\n 一、关于引用文獻一覽表之記載\n 審查人員制作新型專利技術報告時，選取適用之全部先前技術文獻相關資料，彙總記載于"引用文獻一覽表"欄位內。\n 二、关于引用文獻之記載\n（一）每一請求項下之引用文獻，係針對該請求項所適用之文獻，僅須記載該文獻的序號。每一請求項之引用文獻多寡不一，撰寫方式可為"引用文獻：1"或"引用文獻：1 及 3"。\n（二）適用情形如下；\n1. 否定新穎性等要件，應記載与請求項之記載內容最接近或最適当之先前技術文獻。\n2. 无法否定新穎性等要件，應記載該技術領域中一般技術水准之參考文獻（代碼 6）。<br>After comparing each claim item in the patent application scope of the new patent technical report, the records of the cited documents are explained as follows: \n1. Records on the list of cited documents\nWhen the examiner prepares the new patent technical report, he selects all applicable previous technical documents and records them in the "list of cited documents" column. \n2. Records on cited documents\n(1) The cited documents under each claim item are the documents applicable to the claim item, and only the serial number of the document needs to be recorded. The number of cited documents for each claim item varies, and the writing method can be "cited document: 1" or "cited document: 1 and 3". \n(2) The applicable situations are as follows; \n1. To negate the requirements of novelty, the previous technical documents that are closest or most appropriate to the content recorded in the claim item should be recorded. \n2. If the requirements of novelty cannot be negated, the reference documents of the general technical level in the technical field should be recorded (code 6). |